# DEEP MULTIMODAL LEARNING FOR EMOTION RECOGNITION IN SPOKEN LANGUAGE

*Yue Gu, Shuhong Chen, Ivan Marsic*

Department of Electrical and Computer Engineering,
Rutgers University, Piscataway, NJ, USA
{yue.guapp, sc1624, marsic}@rutgers.edu

## ABSTRACT

In this paper, we present a novel deep multimodal framework to predict human emotions based on sentence-level spoken language. Our architecture has two distinctive characteristics. First, it extracts the high-level features from both text and audio via a hybrid deep multimodal structure, which considers the spatial information from text, temporal information from audio, and high-level associations from low-level handcrafted features. Second, we fuse all features by using a three-layer deep neural network to learn the correlations across modalities and train the feature extraction and fusion modules together, allowing optimal global fine-tuning of the entire structure. We evaluated the proposed framework on the IEMOCAP dataset. Our result shows promising performance, achieving 60.4% in weighted accuracy for five emotion categories.

## 1. INTRODUCTION

Human speech conveys both content and attitude. When communicating through speech, humans naturally pick up both content and emotions to understand the speaker's actual intended meaning. Emotion recognition, defined as extracting a group of affective states from humans, is necessary to automatically detect human meaning in a human-computer interaction. Speech emotion recognition, under the field of affective computing, extracts the affective states from speech and reveals the attitudes under spoken language.

Compared to the large amount of research in visual-audio multimodal emotion recognition, there is relatively little work combining text and audio modalities. To detect the emotions in utterances, humans often consider both the textual meaning and prosody. A multimodal structure is thus necessary for using both the text and audio as input data. Previous research shows promising performance improvements by combining text with acoustic information, demonstrating the potential benefits of textual-acoustic structures [1, 2]. One challenge to successfully recognizing human emotions is the extraction of effective features from speech data. There are a number of widely used low-level handcrafted features used for sentiment analysis and emotion detection in natural language and speech signal processing. In particular, thousands of low-level acoustic descriptors and derivations (LLD) with functional statistics are extracted via OpenSmile software in [2, 3]; bag of words (BoW) and bag of n-grams (BoNG) were extracted from text to represent linguistic features [4, 5, 6]. Nevertheless, these low-level features poorly represent high-level associations and are considered insufficient to distinguish emotion [1, 2, 7, 8]. In [1, 2], a convolutional neural network (ConvNet) extracted the high-level textual features from word embedding maps to represent textual features; however, they still combined it with handcrafted low-level acoustic features in the shared representation. Although ConvNets can extract high-level acoustic features [9, 10], they do so without considering the temporal associations. Hence, a common structure that extracts high-level features from both text and audio is desirable.

Another challenge in emotion recognition is the fusion of different modalities. There are two major fusion strategies for multimodal emotion recognition: decision-level fusion and feature-level fusion. Unlike decision-level fusion that combines the unimodal results via specific rules, feature-level fusion merges the individual feature representations before the decision making, significantly improving performance [5, 6], especially in recent deep models [1, 2, 11]. Nevertheless, these works directly feed the concatenated features into a classifier or use shallow-layered fusion models, which have difficulty learning the complicated mutual correlations between different modalities. A deep belief network that consists of three Restricted Boltzmann Machine layers achieves better performance than shallow fusion models by fusing the high-level audio-visual features [12]; however, it separates the training stage of feature extraction and feature fusion. The biggest issue with this approach is that it cannot guarantee global tuning of the parameters, as the prediction

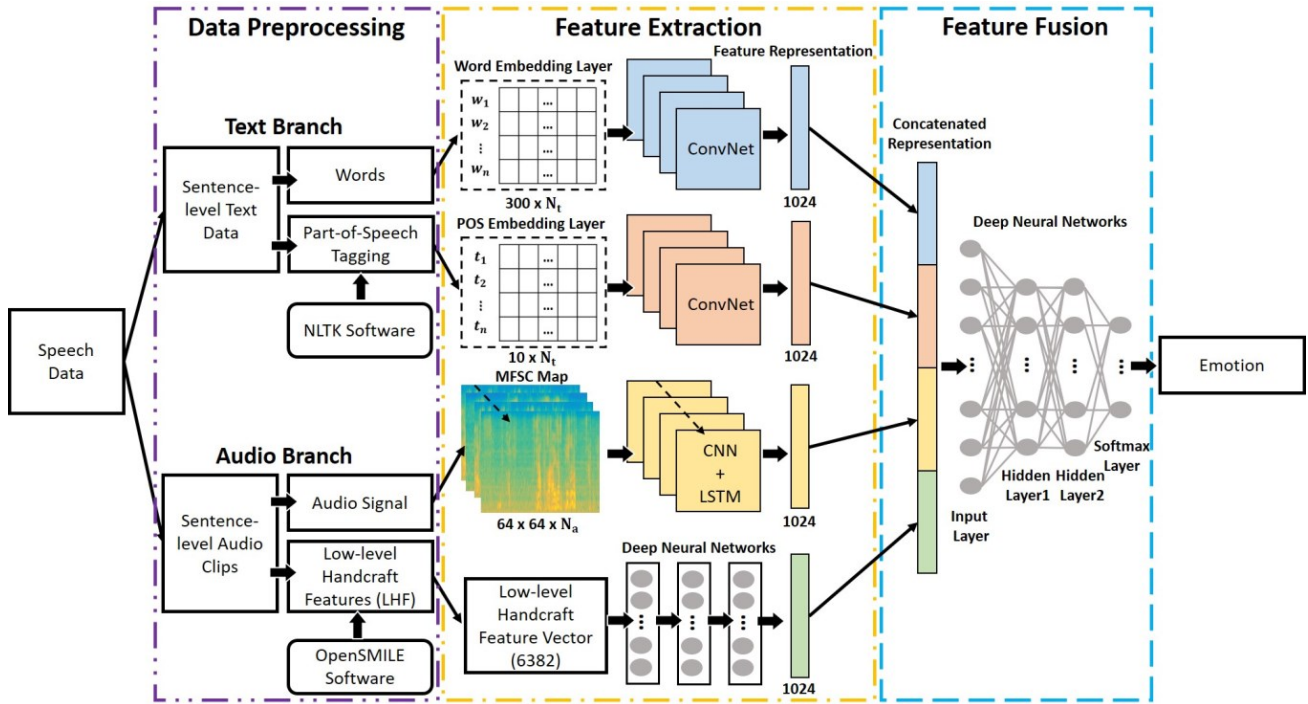

Fig.1. Overall structure of the proposed deep multimodal framework

loss is not actually backpropagated to tune the feature extraction module.

In this paper, we propose a deep multimodal framework to address the problems above. To predict human emotions from sentence-level spoken language, we build a hybrid deep model structure. It uses ConvNets to extract textual features from words and part-of-speech, a CNN-LSTM structure to capture spatial-temporal acoustic features from Mel-frequency spectral coefficients (MFSCs) energy maps, and a three-layer deep neural network to learn high-level acoustic associations from low-level handcrafted features. We then concatenate all the extracted features by using a three-layer deep neural network to learn the mutual correlations across modalities and classify the emotions via a softmax classifier. We directly train the feature extraction module and fusion model together, so that the final loss is appropriately used to tune all parameters. The proposed structure achieves 60.4% weighted accuracy for five emotions on the IEMOCAP multimodal dataset. We also demonstrate the promising performance compared with previous multimodal structures.

## 2. PROPOSED METHOD

As shown in Fig.1, The proposed deep multimodal framework consists of three modules: data preprocessing, feature extraction, and feature fusion. The data preprocessing module processes the input speech streams and outputs the corresponding text sentence, part-of-speech tags, audio signal, and extracted low-level handcrafted acoustic features. Then, a hybrid deep structure initializes and extracts the textual and acoustic features from the above four input branches, respectively. The fusion module concatenates the output features as a joint feature representation and learns the mutual correlations through a deep neural network. We use a softmax layer to finally predict the emotions based on the final shared representation.

### 2.1. Data Preprocessing

We first divide the input speech streams into sentence-level text and the corresponding audio clips. We used Natural Language Toolkit (NLTK) to extract the part-of-speech tags (POS) for each sentence to help to identify the human speaking manner [13]. We remove all the punctuation in both the text and POS. Instead of just using audio signals as input data (spectral feature maps from the feature extraction module), we also extract the low-level pitch and vocal related features using OpenSmile software [14]. Specifically, the software extracts low-level descriptions such as fundamental frequency, pitch/energy related features, zero crossing rate (ZCR), jitter, shimmer, mel-frequency cepstral coefficients (MFCC), etc., with some functional statistics, such as flatness, skewness, quartiles, standard deviation, root quadratic mean, etc. The total number of the features is 6382. As shown in Fig 1, we feed all the four branches into the feature extraction module.

### 2.2. Feature Extraction

To initialize the words, we first use *word2vec* (a pre-trained word embedding model with 300 dimensions for each word based on 100 million words from Google news [15]) as a dictionary to embed each word into a low-dimensional word vector. We pad all sentences with zero padding to fit 40×300. As suggested in [16], we apply one convolutional layer with one max-pooling layer to extract the features and use multiple convolutional filters with 2, 3, 4, and 5 as the widths. We created 256 filters for each width. The final textual feature representation is a 1024-dimensional feature vector.

For POS embedding, we did not use a pre-trained dictionary as we did with word embedding; instead, we trained our own POS embedding dictionary based on the *word2vec* model using our own POS tagging data. We encoded the POS into a 10-dimensional vector and used the same ConvNet structure as the word branch to extract the POS features. We also created 256 filters for each width and made the output POS feature representation a 1024-dimensional feature vector.

| MFSC Map | Conv (3×3×32) Max-pool (2×2) | Conv (3×3×64) Max-pool (2×2) | Conv (3×3×128) Max-pool (2×2) | Conv (3×3×256) Max-pool (2×2) | Fully-Connected Layer (4096) | Dense Layer (1024) | LSTM (1024) |
|---|---|---|---|---|---|---|---|
| | Layer1 | Layer2 | Layer3 | Layer4 | Layer5 | Layer6 | Layer7 |

Fig.2. Feature extraction structure for MFSC maps.

For the audio signal input, we first extracted Mel-frequency spectral coefficients (MFSCs) from raw audio signals, which were shown to be efficient in convolutional models of speech recognition and intention classification in recent study [11, 17, 18]. Compared to the MFCCs, MFSCs maintain the locality of the data by preventing new basis of spectral energies resulting from discrete cosine transform in MFCC extraction [17]. We used 64 filter banks to extract the MFSCs and extracted both the delta and double delta coefficients. Instead of resizing the MFSC feature maps into the same size as in [18], we selected 64 as the context window size and 15 frames as the shift window to segment the entire MFSC map. In particular, given an audio clip, our MFSC map is a 4D array with size n×64×64×3, where n is the number of shift windows. We constructed an eight-layer ConvNet to capture the spatial associations from each MFSC segmentation, which has four convolutional layers with four max-pooling layers. As shown in Fig.2, we selected 3×3 as the convolutional kernel size and 2×2 as the max-pooling kernel size. We applied a fully-connected layer and a dense layer to connect feature vectors. Although previous research used a 3D-CNN structure to learn the temporal associations from the spectrograms [12], simply concatenating output features from the ConvNet cannot reveal the actual temporal associations in sequence. LSTM is a special recurrent neural network (RNN) that allows input data with varying length, remembers values with arbitrary intervals, learns the long-term dependencies of time series, and outputs a fixed-length result. Compared with the ConvNet, LSTM is more suitable to capture the temporal associations, as it considers the sequential properties of the time series. We set up an LSTM layer after the dense layer (Layer6) to handle segmented sequential output with various lengths and learn temporal associations. We selected the hidden state from the last layer (Layer7) as the final 1024-dimensional feature vector output.

Despite the high-level acoustic features from spectral energy maps, we also extract the low-level features in prosody and vocal quality. Unlike most previous research that concatenated the low-level handcrafted features directly or reduced the dimension of the feature vectors via correlation-based feature selection (CFS) and principle component analysis (PCA) [1, 2], we set up a three-layer deep neural network of one input layer with two hidden layers to extract the high-level associations from the low-level features. Max-min normalization is applied for the low-level features before feeding them into the network. The input layer is a 6382-dimensional feature vector and we set 2048 and 1024 as the hidden units for each hidden layer, respectively. We select the last hidden layer as the final feature representation, which is a 1024-feature vector.

### 2.3. Feature Fusion

We concatenate all the extracted high-level features to form the joint feature representation. We use a deep neural network with one input layer, two hidden layers, and a softmax layer to capture the associations between the features from different modalities and classify the emotions. The hidden units are 2048 and 1024 for each hidden layer, respectively. The output of the softmax layer is the corresponding emotion vector. It worth mentioning that we also try to replace the softmax function with a linear SVM to classify the shared representation from the last hidden layer in the fusion model. Nevertheless, there is no obvious improvement in performance. To eliminate the unnecessary structures, we directly use softmax as the final classifier.

### 2.4. Network Training

Unlike previous research that trained the feature extraction module and fusion modules separately, our architecture connects them together and uses backpropagation to adjust the entire framework, including the parameters in both fusion and feature extraction modules. Considering the multiple layers in the proposed structure, we use the rectified linear unit (ReLU) as the activation function to facilitate convergence and set dropout functions to overcome overfitting. Another issue for training a deep model is internal covariate shift, which is defined as the change in the distribution of network activations due to the change in network parameters during training [19]. We applied batch normalization function between each layer to normalize and better learn the distribution [19], improving the training efficiency. We initialize the learning rate at 0.01 and use Adam optimizer to minimize the value from categorical cross-entropy loss function.

### 3. EXPERIMENT AND EVALUATION

We evaluate our proposed framework on the Interactive Emotional Dyadic Motion Capture Database (IEMOCAP) [20]. IEMOCAP is a multimodal emotion dataset including visual, audio, and text data. In this research, we only consider the audio and text data. Three annotators assign one emotion label to each sentence from *happy, sad, neutral, anger, surprised, excited, frustration, disgust, fear,* and *other*. We only use the sentences with at least two agreed emotion labels for our experiments. Followed by the previous research [2], we merged *excited* and *happy* as *Hap*, making the final dataset 1213 *Hap,* 1032 *Sad (sad),* 1084 *Ang (anger),* 774 *Neu (neutral),* and 1136 *Fru (frustration).* We apply 5-fold cross validation to train and test the framework.

**Table 1**. Accuracy comparison of different feature combinations (percentage)

| *Approach* | *Ang* | *Hap* | *Sad* | *Neu* | *Fru* |
|---|---|---|---|---|---|
| $CNN_{word}$ | 42.9 | 54.0 | 50.2 | 39.7 | 49.2 |
| $CNN_{pos}$ | 10.3 | 33.2 | 30.3 | 12.9 | 39.5 |
| $CNN\_LSTM_{mfsc}$ | 51.5 | 50.6 | 52.3 | 43.2 | 49.2 |
| $DNN_{lhaf}$ | 54.3 | 44.1 | 40.4 | 39.8 | 41.7 |
| $CNN_{word} + CNN_{pos}$ | 47.5 | 54.1 | 53.3 | 41.5 | 49.3 |
| $CNN_{word}+CNN\_LSTM_{mfsc}$ | 54.6 | 59.2 | 57.2 | 52.1 | 54.3 |
| $CNN_{word}+DNN_{lhaf}$ | 55.3 | 52.5 | 54.2 | 51.2 | 52.2 |
| $CNN_{pos}+CNN\_LSTM_{mfsc}$ | 46.1 | 40.3 | 41.3 | 34.2 | 40.4 |
| $CNN_{pos}+DNN_{lhaf}$ | 37.2 | 42.8 | 35.3 | 27.7 | 35.4 |
| $CNN\_LSTM_{mfsc}+DNN_{lhaf}$ | 53.7 | 51.3 | 51.1 | 41.3 | 49.5 |
| Both_text+$CNN\_LSTM_{mfsc}$ | 55.7 | 61.3 | 57.4 | 52.6 | 57.5 |
| Both_text+$DNN_{lhaf}$ | 55.9 | 60.2 | 54.1 | 50.3 | 54.3 |
| $CNN_{word}$+Both_audio | 56.1 | 63.2 | 60.1 | 55.4 | 60.4 |
| $CNN_{pos}$+Both_audio | 47.2 | 42.3 | 40.1 | 36.2 | 40.5 |
| Our Method_Separate | 55.3 | 61.4 | 57.2 | 52.3 | 58.1 |
| **Our Method _Together** | **57.2** | **65.8** | **60.2** | **56.3** | **61.6** |

$CNN_{word}$: Using ConvNet as feature extractor and text as input; $CNN_{pos}$: Using ConvNet as feature extractor and part-of-speech tags as input data; $CNN\_LSTM_{mfsc}$: Using CNN-LSTM as feature extractor and MFSC energy maps as input data; $DNN_{lhaf}$: Using DNN as feature extractor and low-level handcraft features as input data; *Both_text*: Including both $CNN_{word}$ and $CNN_{pos}$; *Both_audio*: Including both $CNN\_LSTM_{mfsc}$ and $DNN_{lhaf}$.

We first evaluate each feature branch individually. As shown in Table 1, the $CNN_{word}$ has good performance on *Sad* and *Hap* category. Compared to high-level acoustic features extracted from low-level handcrafted features ($DNN_{lhaf}$), the spatial-temporal high-level acoustic features extracted from the CNN-LSTM lead to better performance on *Hap, Sad, Neu,* and *Fru. $DNN_{lhaf}$* achieves the best result on *Ang* category in all unimodal structures, with 54.3% accuracy. Then, we compare the performance of different feature combinations. Combining all the features from four branches achieves the best result, with 60.4% weighted accuracy. We evaluate different training manners: training the feature extraction module and fusion module separately (Our Method _Separate), and training all modules together (Our Method_Together). Our result shows that training the entire structure together increases weighted accuracy by 2.7%.

We also conducted experiments using methods proposed in previous research. From Table 2, our framework outperforms the text-specific model ($BoW$ and $CNN_{word}$) and acoustic-specific model ($LHAF_w$ and $CNN_{mel}$) by 9.9%-29.5% accuracy. Compared with the low-level textual features ($BoW$), high-level textual features ($CNN_{word}$) improve the accuracy around 6% on average. The high-level acoustic features extracted from Mel-spectrogram via ConvNet structure ($CNN_{mel}$) perform slightly better than the low-level handcrafted acoustic features without feature selection ($LHAF_{wo}$). From our result, using PCA and CFS to select the low-level handcrafted acoustic features ($LHAF_w$) helps improve performance less. Both $LHAF_{wo}$ and $LHAF_w$ have lower weighted accuracies compared to $DNN_{lhaf}$ in Table 1. We also evaluate structures using shallow layers in the fusion model [2, 11]; our proposed hybrid deep multimodal structure achieves the best performance, improving accuracy by up to 8%. It is worth noting that simply replacing the low-level handcrafted features with high-level features from $CNN_{mfsc}$ in the multimodal structure does not significantly improve performance. Using $CNN\_LSTM_{mfsc}$ as the feature extractor improves 3.9% weighted accuracy, demonstrating that the lack of temporal associations indeed influences system accuracy. Our experiments also show that using a linear SVM as the classifier after the deep model does not significantly improve performance compared to a single softmax classifier.

**Table 2**. Comparision of previous emotion recognition structures (percentage)

| *Approach* | *Ang* | *Hap* | *Sad* | *Neu* | *Fru* |
|---|---|---|---|---|---|
| $BoW$+SVM | 40.6 | 45.0 | 42.2 | 31.7 | 44.2 |
| $CNN_{word}$[16] | 42.9 | 54.2 | 50.3 | 39.7 | 49.2 |
| $LHAF_{wo}$+SVM [1] | 41.2 | 36.6 | 38.3 | 39.2 | 41.5 |
| $LHAF_w$+SVM [1] | 40.2 | 37.1 | 40.2 | 40.1 | 41.8 |
| $CNN_{mel}$ [7] | 39.7 | 41.2 | 43.5 | 39.1 | 41.4 |
| $CNN_{word}$+$LHAF_w$+MKL[2] | 50.3 | 52.5 | 53.2 | 49.2 | 52.2 |
| $CNN_{word}$+ $CNN_{mfsc}$ [11] | 50.1 | 52.3 | 56.3 | 51.2 | 50.4 |
| $CNN_{word}$+$CNN_{mfsc}$+SVM | 51.2 | 50.8 | 55.3 | 51.7 | 51.4 |
| **Our Method** | **57.2** | **65.8** | **60.2** | **56.3** | **61.6** |

$LHAF_{wo}$: Low-level handcrafted acoustic features without feature selection. $LHAF_w$: Low-level handcraft acoustic features with feature selection. $CNN_{mel}$: Using ConvNet as feature extractor and mel-spectrogram as input data. $CNN_{mfsc}$: Using ConvNet as feature extractor and MFSC as input data. *MKL*: Using multiple kernel learning as fusion strategy.

## 4. CONCLUSION

In conclusion, we proposed a hybrid deep framework to predict the emotions from spoken language, which consists of ConvNets, CNN-LSTM, and DNN, to extract spatial and temporal associations from the raw text-audio data and low-level acoustic features. We used a four-layer deep neural network to fuse the features and classify the emotions. Our results show that the proposed framework outperforms the previous multimodal structures on the IEMOCAP dataset, achieving 60.4% weighted accuracy on five emotion categories.

## ACKNOLEDGEMENTS
We would like to thank reviewers for the valuable feedback and the SAIL-USC for providing us the IEMOCAP dataset.